\documentclass[letterpaper, 10 pt, conference]{ieeeconf}  

\IEEEoverridecommandlockouts                              

\usepackage{amsthm}
\usepackage{mathtools}
\usepackage{xcolor}
\usepackage{layouts}
\usepackage{graphicx} 
\usepackage{amsmath} 
\usepackage{amssymb}  
\usepackage{tikz}
\usepackage{multirow}

\usepackage[hyphens]{url}
\usepackage[hidelinks]{hyperref}
\usepackage{hyperref}
\usepackage{cite}
\usepackage{pgf}

\usepackage{hyperref}

\newcommand{\footref}[1]{%
    $^{\ref{#1}}$%
}

\usepackage{bm}
\usepackage{array}
\usepackage{booktabs}

\usepackage{tikz-3dplot}
\usepackage{lipsum}
\usepackage{subcaption}
\usepackage[font=footnotesize,labelfont=bf]{caption}

\newtheorem{theorem}{Theorem}

\makeatletter
\renewcommand*\env@matrix[1][\arraystretch]{%
 \edef\arraystretch{#1}%
 \hskip -\arraycolsep
 \let\@ifnextchar\new@ifnextchar
 \array{*\c@MaxMatrixCols c}}
\makeatother

\title{\LARGE \bf
Online Spatio-temporal Calibration of Tightly-coupled Ultrawideband-aided Inertial Localization
}

\author{Abhishek Goudar and Angela P. Schoellig
\thanks{The authors are with the Dynamic Systems Lab, Institute for Aerospace Studies, University of Toronto, Canada, and affiliated with the Vector Institute for Artificial Intelligence in Toronto. E-mails: {\tt \{firstname.lastname\}@robotics.utias.utoronto.ca}}%
}

\begin{document}

\maketitle
\thispagestyle{empty}
\pagestyle{empty}

\begin{abstract}
The combination of ultrawideband (UWB) radios and inertial measurement units (IMU) can provide accurate positioning in environments where the Global Positioning System (GPS) service is either unavailable or has unsatisfactory performance. The two sensors, IMU and UWB radio, are often not co-located on a moving system. The UWB radio is typically located at the extremities of the system to ensure reliable communication, whereas the IMUs are located closer to its center of
gravity. Furthermore, without hardware or software synchronization, data from heterogeneous sensors can arrive at different time instants resulting in temporal offsets. If uncalibrated, these spatial and temporal offsets can degrade the positioning performance. In this paper, using \emph{observability} and \emph{identifiability} criteria, we derive the conditions required for successfully calibrating the spatial and the temporal offset parameters of a \emph{tightly-coupled} UWB-IMU system. We also present an \emph{online} method for \emph{jointly} calibrating these offsets. The results show that our calibration approach results in improved positioning accuracy while simultaneously estimating \emph{(i)} the spatial offset parameters to millimeter precision and \emph{(ii)} the temporal offset parameter to millisecond precision.
\end{abstract}

\section{INTRODUCTION}

Global Positioning System (GPS) is the \textit{de facto} standard for positioning in outdoor environments. However, an equivalently powerful localization scheme for indoor environments is missing and GPS performance can be significantly degraded in urban canyons. With the shrinking costs of high-quality inertial measurement units (IMUs) and cameras, camera-IMU-based localization systems have gained popularity for GPS-denied navigation. Yet, the reliable operation of such systems requires good illumination, persistent and distinguishable features, loop closures, and the dominant part of the scene to be static and occlusion-free. In highly dynamic environments such as warehouses, factories, hospitals and shopping malls, these requirements can be restrictive.

%
\begin{figure}[h]
	\centering
	\tdplotsetmaincoords{70}{110}
	\begin{tikzpicture}[scale=1.3,tdplot_main_coords] 
		\coordinate (WORLD_ORIGIN) at (0,0,0);
		\coordinate (UWB_ORIGIN) at (3,3.5,3.6);
		\coordinate (IMU_ORIGIN) at (3,5.2,4.1);
		\coordinate (ANCHOR_ORIGIN) at (0,1,4);
		\coordinate (RADIO_ORIGIN) at (0,1,3.85);
		\coordinate (DRONE_ORIGIN) at (2.5,5,3.6);

		\node [anchor=north] at (WORLD_ORIGIN){$\{W\}$};
		\node [anchor=north] at (IMU_ORIGIN){$\{I\}$};
		\node [anchor=north east] at (UWB_ORIGIN){$\{U\}$};
		\node [anchor=north east] at (ANCHOR_ORIGIN){$\{A_j\}$};

		\draw[solid] (UWB_ORIGIN) -- (IMU_ORIGIN) node[above, pos=0.45]{$\textbf{p}^I_U$};
		\draw[dashed] (WORLD_ORIGIN) -- (IMU_ORIGIN) node[below right, pos=0.4]{$\{\textbf{p}^W_I, \textbf{q}^W_I\}$};
		\draw[dashed] (WORLD_ORIGIN) -- (ANCHOR_ORIGIN) node[above left, pos=0.5]{$\textbf{p}^W_j$};
		\draw[dotted] (ANCHOR_ORIGIN) -- (UWB_ORIGIN) node[above right, pos=0.45]{$r_j$};

		\draw[thick,->] (WORLD_ORIGIN) -- (0,1,0) node[anchor=north west]{$x$};
		\draw[thick,->] (WORLD_ORIGIN) -- (-1.5,0,0) node[anchor=south west]{$y$};
		\draw[thick,->] (WORLD_ORIGIN) -- (0,0,1) node[anchor=south]{$z$};

		\tdplotsetrotatedcoords{0}{0}{0}
		\tdplotsetrotatedcoordsorigin{(UWB_ORIGIN)}\draw[thick,color=black,tdplot_rotated_coords,->] (0,0,0) --(-0.7,0,0) node[anchor=south]{$x$};
		\draw[thick,color=black,tdplot_rotated_coords,->] (0,0,0) --(0,-0.5,0) node[anchor=south]{$y$};
		\draw[thick,color=black,tdplot_rotated_coords,->] (0,0,0) --(0,0,0.5) node[anchor=south]{$z$};

		\tdplotsetrotatedcoords{0}{0}{0}
		\tdplotsetrotatedcoordsorigin{(IMU_ORIGIN)}\draw[thick,color=black,tdplot_rotated_coords,->] (0,0,0) --(-.7,0,0) node[anchor=south]{$x$};
		\draw[thick,color=black,tdplot_rotated_coords,->] (0,0,0) --(0,-0.5,0) node[anchor=south]{$y$};
		\draw[thick,color=black,tdplot_rotated_coords,->] (0,0,0) --(0,0,0.5) node[anchor=south]{$z$};

		\tdplotsetrotatedcoords{0}{0}{0}
		\tdplotsetrotatedcoordsorigin{(ANCHOR_ORIGIN)}\draw[thick,color=black,tdplot_rotated_coords,->] (0,0,0) --(-.7,0,0) node[anchor=south]{$x$};
		\draw[thick,color=black,tdplot_rotated_coords,->] (0,0,0) --(0,-0.5,0) node[anchor=south]{$y$};
		\draw[thick,color=black,tdplot_rotated_coords,->] (0,0,0) --(0,0,0.5) node[anchor=south]{$z$};
		
		\node[inner sep=0pt, opacity=0.25] (drone) at (DRONE_ORIGIN) {\includegraphics[width=.30\textwidth]{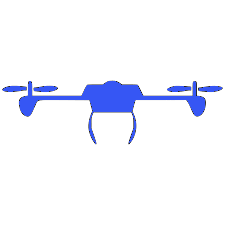}};
		\node[inner sep=0pt, opacity=0.5] (mobradio) at (UWB_ORIGIN) {\includegraphics[width=.015\textwidth]{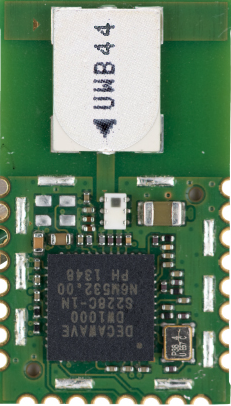}};
		\node[inner sep=0pt, opacity=0.5] (anchor) at (RADIO_ORIGIN) {\includegraphics[width=.02\textwidth]{figs/dw1000.png}};
	\end{tikzpicture}
	\caption{The setup considered consists of an IMU (frame $\{I\}$) and an UWB radio (frame $\{U\}$) mounted in a non-collocated manner on a mobile robot. Localization is performed by measuring the distance between the radio and the UWB anchors. Frame $\{W\}$ corresponds to a gravity-aligned world reference frame. The position of the $j^{th}$ anchor (frame $\{A_j\}$) in the world frame is $\textbf{p}^W_j$ and the measured distance between the radio and the $j^{th}$ anchor is denoted by $r_j$. The pose of the IMU in the world frame is {$\{\textbf{p}^W_I, \textbf{q}^W_I\}$}. The spatial offset, $\mathbf{p}^I_U$, is the position of the radio in the IMU frame.}
	\label{fig:setup}
\end{figure}

Ultrawideband (UWB) radio technology is a promising sensor alternative; it is immune to illumination changes, it does not require loop closures or persistent features, and it can operate in visually challenging conditions. As such, there has been extensive work on UWB-aided localization \cite{Hol2009, Prorok2014, Mueller2015}. A typical UWB-based positioning system consists of radios, known as \textit{anchors}, installed in the periphery of an area of interest, as shown in Fig. \ref{fig:setup}. A mobile agent equipped with a UWB radio calculates its position by measuring the time-of-flight (ToF) between its radio and the anchors. This setup is similar to GPS, where a receiver calculates its position by measuring the ToF to multiple satellites. However, UWB-based positioning systems have several advantages: they are portable, easy to install, rely on less expensive infrastructure, and provide higher accuracy (2-10\,cm).

We refer to localization solutions that \emph{(i)} compute position from range measurements first (as done in GPS) and then \emph{(ii)} use that value in a state estimator as \emph{loosely-coupled}. Such a system can be restrictive; for example, for GPS, no position estimate is available whenever less than four satellites are visible \cite{kaplan2005}. We refer to a system that directly uses the range measurements as inputs to a state estimator as \emph{tightly-coupled}. Such a system can update its internal state even when a single anchor is available. UWB radios can be used in two-way time-of-flight (TW-ToF) mode---in which the range measurements between an anchor and a radio are obtained by sending a signal from the radio to the anchor and back. In this paper, we implement a tightly-coupled UWB-based localization system which operates by the measuring the TW-ToF to one anchor at a time.

A mobile system equipped with a UWB radio, however, can estimate only 3D position. To obtain a 6 degree-of-freedom pose, IMUs are used in conjunction~\cite{Hol2009}. Generally, IMUs and UWB radios are not co-located and there is a spatial offset between the two sensors, also referred to as \emph{sensor extrinsic parameters}. Current calibration techniques involve measuring the spatial offset manually, using survey equipment or additional sensors. These methods are prone to error and expensive. While computer-aided diagrams provide the mechanical specifications of a radio, the exact position of the \emph{phase center}\footnote{In antenna design theory, the phase center is the point from which the electromagnetic radiation spreads spherically outward.} is generally not known.

\begin{figure}[!t]
   \centering
    \includegraphics[width=0.8\linewidth]{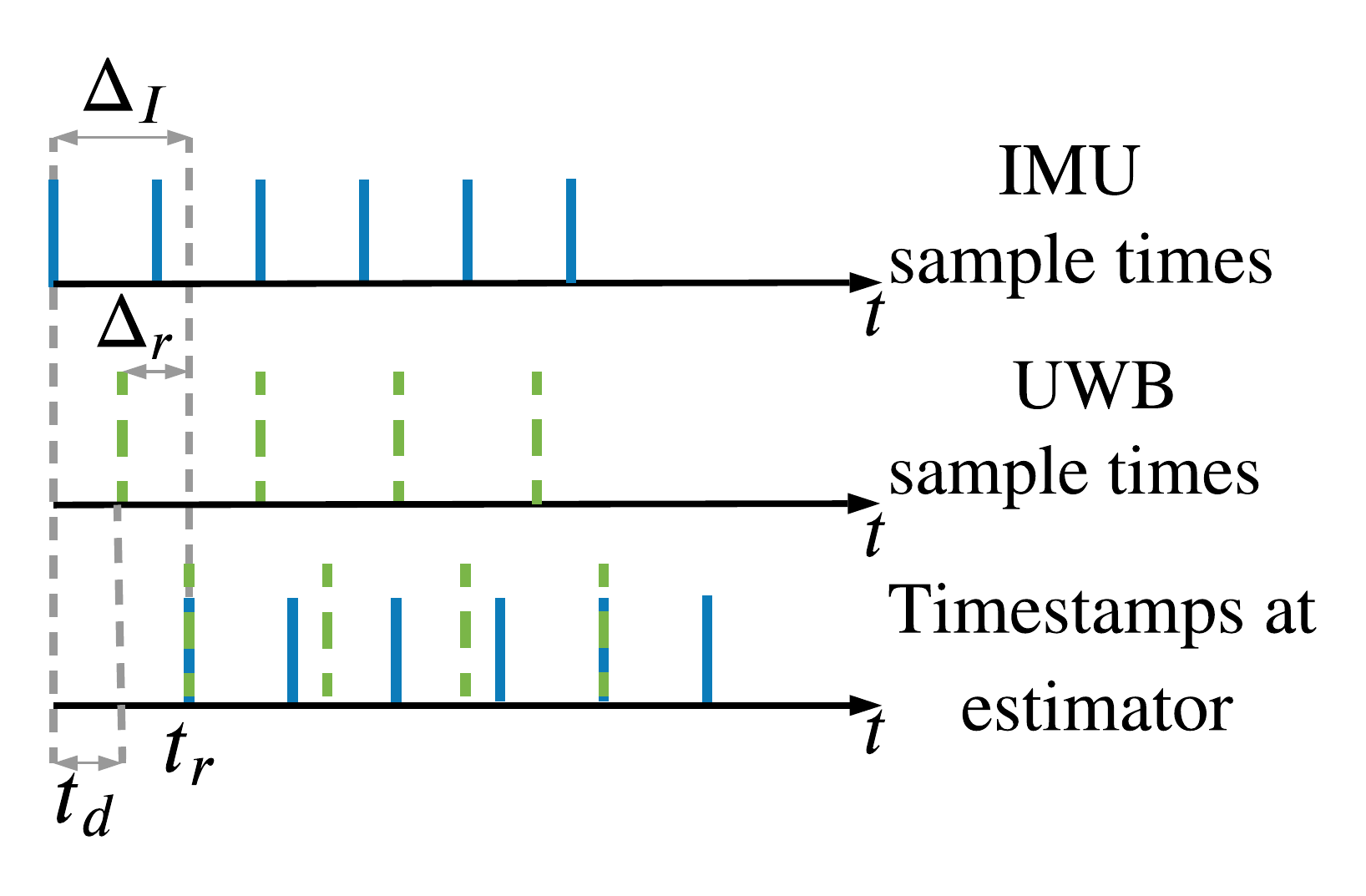}
    \caption{Sensor latency in the UWB radio ($\Delta_r$) and the IMU ($\Delta_I$) causes data generated at the different time instants to arrive at the same time $t_r$ at the estimator. By estimating the temporal offset $t_d = \Delta_I - \Delta_r$, UWB and IMU data can be realigned on a common time scale.}
    \label{fig:temp_calib_mot}
\end{figure}

Our setup requires the IMU and the UWB data to have accurate timestamps with respect to a single source of clock. This is generally achieved through hardware synchronization using a common clock signal. The next best choice is software synchronization with a clock server running on a destination computer and a client on the sensor hardware. However, many off-the-shelf components do not support either of these methods. Hence, due to sensor latency and different clock sources, there is a \emph{temporal} offset between data from heterogeneous sensors as shown in Fig.~\ref{fig:temp_calib_mot}. Estimating the state without compensating for the spatio-temporal offsets can result in poor positioning accuracy, particularly when these offsets are large. In this paper, we propose an online calibration procedure that uses the available sensors only for estimating the spatial and the temporal offsets.

An important aspect of calibration is the \textit{observability} and \emph{identifiability} of the relevant states. Observability and identifiability are measures to determine if the internal states of a system can be inferred from its external output (i.e., measurements). In this paper, we perform an observability analysis to derive conditions for the \textit{local weak observability} \cite{Hermann1977} of the internal states and the spatial offset. Using a related property, \emph{local identifiability} \cite{anguelova2008}, we derive the conditions under which the temporal offset is locally identifiable. Observability analysis helps us answer the question of whether a particular system and its state is observable or not. However, it does not lend itself to the design of a suitable estimator, especially in the presence of noise. In this work, we propose an \emph{online} approach, based on the \emph{error state Kalman filter} (ESKF), to jointly estimate the state, and the spatial and temporal offset parameters. In summary, our main contributions are as follows:
\begin{enumerate}
\item we derive the conditions under which the temporal offset is locally identifiable;
\item we derive the conditions under which the state of a tightly-coupled UWB-IMU system, including the spatial sensor offset, is locally weakly observable;
\item we provide a unified approach for estimating the IMU position, orientation, velocity, IMU biases, the spatial offset \emph{and} the temporal offset of a tightly-coupled UWB-IMU system;
\item we evaluate the proposed method on simulated and real data to validate our theory and show the efficacy of the recommended approach. 
\end{enumerate}

\section{Related work}
The use of UWB radios for positioning in indoor and outdoor environments has been demonstrated extensively in the literature. Localization approaches ranging from parametric~\cite{Hol2009, Mueller2015} and non-parametric~\cite{Prorok2011, Prorok2014} to graph-optimization-based~\cite{Fang2018} methods have been proposed. Estimation of the UWB-IMU spatial offset with the aid of GPS and cameras was shown in~\cite{Hausman2016}. However, most of the previous works either assume known spatial offset parameters or calibrate only the spatial offset using additional hardware.

Spatio-temporal calibration of camera-IMU systems has been studied extensively~\cite{Kelly2011, Weiss2012, Hesch2014, Li2014, Furgale2013}. Estimation of IMU pose, velocity, biases, visual landmark positions, scale factor, and camera-IMU spatial and temporal offsets has been conducted in~\cite{Li2014}. However, the measurement of a tightly-coupled UWB-IMU system is a distance measurement and not a pose. The calibration methods proposed for camera-IMU systems are not applicable to the system considered in this paper. 

GPS-based positioning schemes are similarly affected by spatial and temporal offsets. In~\cite{Hong2005}, estimation of the spatial offset for a loosely-coupled GPS-IMU system is achieved with an extended Kalman filter (EKF). Time synchronization error calibration for a loosely-coupled GPS-aided inertial navigation system is shown in~\cite{skog2011}. Each of the previously mentioned works calibrate either the spatial offset or the temporal offset, but not both. Additionally, as alluded to earlier, loosely-coupled systems have limitations. To the best of the authors' knowledge, \emph{joint} spatio-temporal calibration for a tightly-coupled UWB-IMU system has not been done.

The observability of the state is crucial for reliable estimation. Previous research has addressed the observability analysis of various sensor combinations; analysis of the observability of camera-IMU systems, including spatial and temporal offsets, has been shown in~\cite{Kelly2011, Li2014}. Since the observation model of a tightly-coupled UWB-IMU system is different from that of a camera-IMU system, the analysis presented in~\cite{Kelly2011, Li2014} is not directly applicable here. In~\cite{Hong2005}, the observability of a linear approximation of a loosely-coupled GPS-IMU system, with the spatial offset only, is shown. In contrast, we derive conditions for the local weak observability of the system state and the spatial offset of a tightly-coupled UWB-IMU system without any approximations. We also derive the conditions under which the temporal offset of such a system is locally identifiable.

\section{Problem formulation} \label{sec:prob_form}

Consider the setup shown in Fig. \ref{fig:setup} wherein a system equipped with an IMU and a UWB radio localizes itself by measuring the TW-ToF between its radio and the anchors. The following assumptions are made:

\begin{description}
\item[(A1)] The spatial offset and the temporal offset are constant and do not change over time.
\item[(A2)] The UWB radio is considered a point source. Hence, its orientation is not considered.
\end{description}

Under these assumptions, the calibration objectives are to:
\begin{description}
\item[(O1)] formulate the conditions under which the tightly-coupled UWB-IMU system with the spatial offset is \textit{locally weakly observable} and the temporal offset is \emph{locally identifiable};
\item[(O2)] improve positioning accuracy by jointly estimating \emph{(i)} the IMU position, orientation, and velocity, \emph{(ii)} the IMU biases, \emph{(iii)} the UWB-IMU spatial offset and \emph{(iv)} the UWB-IMU temporal offset.
\end{description}

\section{System Modelling}
We define the following coordinate frames for the setup shown in Fig. \ref{fig:setup}:
\begin{enumerate}
\item \textbf{world frame} $\{W\}$, a gravity-aligned absolute reference frame, in which the pose of the IMU and the positions of individual anchors are expressed;
\item \textbf{mobile radio frame} $\{U\}$, a frame affixed to the phase center of the mobile radio antenna;
\item \textbf{IMU frame} $\{ I \}$, a frame corresponding to the IMU body center, in which the body accelerations and angular velocities are measured;
\item \textbf{anchor frame} $\{A_i\}$, a frame affixed to the phase center of the $i^{th}$ anchor.
\end{enumerate}

\subsection{System parameterization}
The system in Fig. \ref{fig:setup} is described by the following 20-dimensional state vector:
\begin{equation}
\mathbf{x}(t) = (\mathbf{p}^W_I(t), \mathbf{v}^W_I(t), \mathbf{q}^W_I(t), \mathbf{b}_a(t), \mathbf{b}_\omega(t), \mathbf{p}^I_U, t_d), \label{eqn:state}
\end{equation}
\noindent where, $\{ \mathbf{p}^W_I(t), \mathbf{v} ^W_I(t), \mathbf{q}^W_I(t) \}$ denote the position, translational velocity and orientation of the IMU frame with respect to the world frame. A unit quaternion parameterization is used for representing orientations. In this paper, we follow the convention: $\mathbf{q} = q_0 + q_x \mathbf{i} + q_y \mathbf{j} + q_z \mathbf{k}$, where $q_0$ is the scalar part and $\mathbf{q}_v = (q_x, q_y, q_z)$ is the vector part. We use quaternions for their singularity-free orientation representation. Accelerometer and gyroscope biases are denoted by $\mathbf{b}_a(t)$ and $\mathbf{b}_g(t)$. The UWB-IMU spatial offset is denoted by $\mathbf{p}^I_U$ and $t_d$ is the temporal offset (see Fig. \ref{fig:temp_calib_mot}).

\subsection{Gyroscope and accelerometer model} \label{sec:imu_model}
The measured angular rate by a triaxial gyroscope $\bm{\omega}_m = (\omega_x, \omega_y, \omega_z)$ is related to the true angular rate $\bm{\omega}_t$ as: $\bm{\omega}_m = \bm{\omega}_t + \mathbf{b}_\omega + \mathbf{n}_\omega$, where $\mathbf{b}_\omega$ is the time-varying bias and $\mathbf{n}_{b\omega}$ is a zero-mean additive white Gaussian noise (AWGN) process with covariance $\mathbf{Q}_\omega$, i.e. $\mathbf{n}_\omega \sim \mathcal{N}(\mathbf{0}, \mathbf{Q}_\omega)$. The bias is modelled as driven by another AWGN process $\mathbf{n}_{b\omega} \sim \mathcal{N}(\mathbf{0}, \mathbf{Q}_{b\omega})$: $\dot{\mathbf{b}}_\omega = \mathbf{n}_{b\omega}$.

The measured linear acceleration by a triaxial accelerometer $\mathbf{a}_m = (a_x, a_y, a_z)$ is related to the true linear acceleration $\mathbf{a}_t$ as: $\mathbf{a}_m = \mathbf{a}_t + \mathbf{b}_a + \mathbf{n}_a$, where $\mathbf{b}_a$ is the time-varying bias and $\mathbf{n}_a$ is an AWGN processes of covariance $\mathbf{Q}_a$, i.e. $\mathbf{n}_a \sim \mathcal{N}(\mathbf{0}, \mathbf{Q}_a)$. The bias is driven by another AWGN noise process $\mathbf{n}_{ba} \sim \mathcal{N}(\mathbf{0}, \mathbf{Q}_{ba})$: $\dot{\mathbf{b}}_a = \mathbf{n}_{ba}$.

\subsection{Motion model}
The motion model in this work is a 3D kinematic motion model, where the accelerometer and gyroscope measurements are used as control inputs:
\begin{align}
    &\dot{\mathbf{p}}^W_I = \mathbf{v}^W_I,&
    &\dot{\mathbf{q}}^W_I = \dfrac{1}{2}\bm{\Omega}(\bm{\omega}_t)\mathbf{q}^W_I,\label{eqn:non_linear_model_1}\\
    &\dot{\mathbf{v}}^W_I = \mathbf{R}^W_I \mathbf{a}_t - \mathbf{g}^W,&
    &\dot{\mathbf{b}}_a = \mathbf{n}_{ba},\label{eqn:non_linear_model_2}\\
    &\dot{\mathbf{b}}_g = \mathbf{n}_{b\omega},&
    &\dot{\mathbf{p}}^I_U = \mathbf{0}_{3}, &\\
    &\dot{t}_d = 0, \label{eqn:non_linear_model_3}
\end{align}
where $
\bm{\Omega}(\bm{\omega}) = \begin{bmatrix}
0 & -\bm{\omega}^T\\
\bm{\omega} & -[\bm{\omega}]_\times
\end{bmatrix}$, $\mathbf{g}^W = [0, 0, 9.8]^T m/s^2$ represents the acceleration due to gravity in the world frame, $\mathbf{R}^W_I := \mathbf{R}\{\mathbf{q}^W_I\}$ is the direction cosine matrix, and $[\cdot]_\times$ denotes the skew-symmetric cross-product matrix. 

\subsection{Observation model} \label{sec:obs_model}
In a tightly-coupled system, the observation model is the distance between an anchor and the mobile radio. The measured distance to the $i^{th}$ anchor at time $t_r$ is a function of the anchor position $\mathbf{p}^W_i$ and the state $\mathbf{x}(t_r)$:
\begin{equation}
    h(\mathbf{p}^W_i, \mathbf{x}(t_r)) = \| \mathbf{p}^W_i - \mathbf{p}^W_U(t_r) \|_2 + n_r(t_r),
    \label{eqn:meas_model}
\end{equation}
where $\mathbf{p}^W_U(t_r) = \mathbf{R}^W_I(t_r) \mathbf{p}^I_U  + \mathbf{p}^W_I (t_r)$ is the position of the mobile radio in the world frame and $\|.\|_2$ denotes the $\ell_2$ norm. The position of the anchor $\mathbf{p}^W_i$ is determined using the procedure outlined in Section \ref{sec:real_exp}. The measurement noise $n_r(t_r)$ is assumed to be a zero-mean AWGN process, with covariance $\text{Q}_r$, i.e. $n_r(t_r) \sim \mathcal{N}(0, \text{Q}_r)$. In this paper, we use UWB timestamps as the reference time. Calculating \eqref{eqn:meas_model} requires the value of the state \eqref{eqn:state} at time $t_r$. Let $t_r$ also be the timestamp of the latest IMU measurement. Note that due to the temporal offset $t_d$, the IMU measurement is actually generated at time $t_I = t_r - t_d$ with respect to the UWB timestamp (see Fig. \ref{fig:temp_calib_mot}). To calculate the state at time $t_r$, we propagate the state at time $t_I$ for $t_d$ seconds using the motion model \eqref{eqn:non_linear_model_1}-\eqref{eqn:non_linear_model_3} and the IMU measurement(s) starting at $t_I$. Note that $t_d$ can be positive or negative. 

\section{Observability and Identifiability analysis} \label{sec:obs_ana}
The goal of the observability analysis is to check if the state of a system can be determined uniquely given the outputs of a system. We decompose the problem of observability of the state~(\ref{eqn:state}) into \emph{(i)} \emph{local identifiability} of $t_d$ \cite{anguelova2008} and \emph{(ii)} \emph{local weak observability} \cite{Hermann1977} of the part of the state excluding $t_d$. The motivation and justification for this two-step process is provided below.

In \cite{anguelova2008}, it is shown that the identifiability of a single unknown constant time-delay can be analyzed independent of the observability of the state. The authors show that the identifiability of time-delay in a nonlinear system is not directly related to the observability of other system states or parameters. Specifically, the time-delay parameter can be determined directly from the \emph{input-output} representation \cite{anguelova2008} of the system, which solely depends on the systems inputs, outputs, and their time-derivatives. Thus, local identifiability of the time-delay is equivalent to determining if an input-output representation exists. Finding such an input-output representation for a general nonlinear system is difficult. However, a necessary and sufficient condition for the existence of an input-output representation for a nonlinear system is the occurrence of the delayed input variables (in our case $\mathbf{a}_m$ and $\bm{\omega}_m$) in the output (\ref{eqn:meas_model}) (Theorem 2 in \cite{anguelova2008}). We use this approach to analyze the local identifiability of $t_d$.

As noted in \cite{anguelova2008}, identifiability of the time-delay does not imply observability of the state. After proving the local identifiability of  $t_d$, we analyze the observability of the part of the state excluding $t_d$, i.e. $\widetilde{\mathbf{x}} = (\mathbf{p}^W_I, \mathbf{v}^W_I, \mathbf{q}^W_I, \mathbf{b}_a, \mathbf{b}_\omega, \mathbf{p}^I_U)$. For this, we use the method outlined in \cite{Hermann1977}, which involves determining the rank of the \emph{observability matrix} $\mathcal{O}$ \cite{Hermann1977}. The methods outlined in \cite{anguelova2008} and \cite{Hermann1977} consider the case of noise-free nonlinear systems. Hence, for this analysis, we neglect the effect of noise parameters in the following sections.

\subsection{Local identifiability of the temporal offset}
Following \cite{anguelova2008}, the local identifiablility of $t_d$ depends on whether it is present in the input-output representation of the system (\ref{eqn:non_linear_model_1})-(\ref{eqn:non_linear_model_3}) and (\ref{eqn:meas_model}), that is, the presence of delayed input variable(s) in the output function. In our case, it is sufficient to show that $\mathbf{a}_m(t_r - t_d)$ or $\bm{\omega}_m(t_r - t_d)$ appear in the measurement model. Without loss of generality, the measurement model (\ref{eqn:meas_model}) can be written as:
\begin{equation}
h(\mathbf{p}^W_i, \mathbf{x}(t_r)) = \frac{1}{2}\| \mathbf{p}^W_i - \mathbf{p}^W_U(t_r) \|^2_2.
\label{eqn:td_meas_model}
\end{equation}
We consider a single anchor as the analysis is identical for multiple anchors. The factor 1/2 is introduced for simplifying the analysis. Since there are two inputs, $t_d$ can be locally identified if $\mathbf{a}_m$ or $\bm{\omega}_m$ is excited. In the following theorem, we state the conditions under which $t_d$ is locally identifiable.

\begin{theorem}
The temporal offset parameter $t_d \in (0, T)$, for some $T \in \mathbb{R}$, is locally identifiable from the observation model (\ref{eqn:td_meas_model}) if:
\begin{description}
\item[(T1)] the mobile radio is not co-located with the anchor; and either {\normalfont{(T2)}} or {\normalfont{(T3)}} is satisfied:
\item[(T2)] at least one of $a_x$, $a_y$, or $a_z$ is excited; or
\item[(T3)] the mobile radio is not co-located with the IMU and all three of $\omega_x$, $\omega_y$, and $\omega_z$ are excited.
\end{description}
\label{lem:td_cond_1}
\end{theorem}
\begin{proof} We provide a proof sketch here. The full proof can be found in the supplementary material\footref{foot:proof}. We approach the proof as follows:
\begin{itemize}
\item[S1.] As alluded to in Section  \ref{sec:obs_model}, we calculate $\mathbf{p}^W_U (t_r)$ by propagating the state \eqref{eqn:state} for $t_d$ seconds using the IMU measurement at time $t_I = t_r - t_d$. We perform an Euler integration step by assuming $\mathbf{a}_m$ and $\bm{\omega}_m$ are constant for $t_d$ seconds. The measurement model \eqref{eqn:td_meas_model} is then given by:
\begin{align}
h(\mathbf{p}^W_i, \mathbf{x}(t_r)) &=\dfrac{1}{2}  \| \mathbf{p}^W_i - g_1(\bm{\omega}_m) 
- g_2(\mathbf{a}_m) \|^2_2, \nonumber
\end{align}
where, 
\begin{align*}
g_1(\bm{\omega}_m) &=  \mathbf{R}^W_I(t_I) \left( \mathbf{p}^I_U  + [\mathbf{p}^I_U]_\times \mathbf{b}_\omega(t_I) t_d \right) \\ &- \mathbf{R}^W_I(t_I) [\mathbf{p}^I_U]_\times \bm{\omega}_m(t_r - t_d) t_d,\\ g_2(\mathbf{a}_m) &= \mathbf{p}^W_I(t_I) + \mathbf{v}^W_I(t_I) t_d \\ 
&+ \frac{1}{2}\mathbf{R}^W_I(t_I)( \mathbf{a}_m(t_r - t_d)  - \mathbf{b}_a(t_I)) t_d^2.
\end{align*}
\item[S2.] We then identify the conditions for which the derivative of the expanded measurement model with respect to the delayed inputs $\mathbf{a}_m(t_r - t_d)$ or $\bm{\omega}_m(t_r - t_d)$ is non-zero:
\begin{align}
\dfrac{\partial h \left( \mathbf{p}^W_i, \mathbf{x}(t_r)\right)}{\partial \mathbf{a}_m(t_r - t_d)} &\ne 0, &
\dfrac{\partial h \left( \mathbf{p}^W_i, \mathbf{x}(t_r) \right)}{\partial \bm{\omega}_m(t_r - t_d)} &\ne 0. \nonumber
\end{align}
\end{itemize}  \qedhere
\end{proof}

Condition T3 is a sufficient condition to handle all pathological cases with a single anchor (see supplementary material\footref{foot:proof}). With multiple non-collinear anchors, excitation of two of $\omega_x$, $\omega_y$, or $\omega_z$ is sufficient. A requirement for local identifiability of $t_d$ is that a change in the input causes a change in the measured range. Condition T3 reflects the fact that if the mobile radio and the IMU are co-located, then the measured range is constant for pure rotational motion.

\subsection{Observability of a tightly-coupled UWB-IMU system}
Now we analyze the observability of $\widetilde{\mathbf{x}}$. The system dynamics (\ref{eqn:non_linear_model_1})-(\ref{eqn:non_linear_model_3}) are rearranged into  a \emph{control affine} form~\cite{Hermann1977}:
\begin{align}
    \dot{\widetilde{\mathbf{x}}} =
    \underbrace{\begin{bmatrix}[1.2]
        \mathbf{v}^W_I\\
        -\mathbf{R}^W_I\mathbf{b}_a - \mathbf{g}^W\\
        -\frac{1}{2}\mathbf{\Xi}\{\mathbf{q}^W_I\}\mathbf{b}_{\omega}\\
        \mathbf{0}_{3}\\
        \mathbf{0}_{3}\\
        \mathbf{0}_{3}\\
        \end{bmatrix}}_{f_0} +
    \underbrace{\begin{bmatrix}[1.2]
        \mathbf{0}_{3 \times 3}\\
        \mathbf{R}^W_I\\
        \mathbf{0}_{3 \times 4}\\
        \mathbf{0}_{3 \times 3}\\
        \mathbf{0}_{3 \times 3}\\
        \mathbf{0}_{3 \times 3}\\
    \end{bmatrix}}_{f_1} \mathbf{a}_m
    + \underbrace{\begin{bmatrix}[1.2]
        \mathbf{0}_{3 \times 3}\\
        \mathbf{0}_{3 \times 3}\\
        \frac{1}{2}\mathbf{\Xi}\{\mathbf{q}^W_I\}\\
        \mathbf{0}_{3 \times 3}\\
        \mathbf{0}_{3 \times 3}\\
        \mathbf{0}_{3 \times 3}\\
    \end{bmatrix}}_{f_2} \bm{\omega}_m,
    \label{eqn:system_dynamics_affine}
\end{align}
with
$
\mathbf{\Xi}\{\mathbf{q}^W_I\} =
    \begin{bmatrix*}[c]
       \multicolumn{1}{c}{-\mathbf{q}_v^T}\\
      q_0 \mathbf{I}_3 -[\mathbf{q}_v]_\times
     \end{bmatrix*}
$. Without loss of generality, the measurement model (\ref{eqn:meas_model}) can be rewritten as:
\begin{equation}
    h(\mathbf{p}^W_i, \widetilde{\mathbf{x}}) = \dfrac{1}{2} \| \mathbf{p}^W_i - \mathbf{R}^W_I \mathbf{p}^I_U - \mathbf{p}^W_I \|^2_2,
    \nonumber
\end{equation}
where we remove the dependency on time for brevity and expand $\mathbf{p}^W_U$ as $\mathbf{p}^W_U = \mathbf{R}^W_I\mathbf{p}^I_U + \mathbf{p}^W_I$. The factor 1/2 is introduced for simplifying the analysis. Measurements from a single anchor are not sufficient to constrain the entire state. Hence, we consider measurements using three anchors $\mathbf{p}^W_a = (\mathbf{p}^W_i,~\mathbf{p}^W_j,~\mathbf{p}^W_k)$:
\begin{equation}
    \mathbf{h}(\mathbf{p}^W_a, \widetilde{\mathbf{x}}) = \dfrac{1}{2}\begin{bmatrix}[1.2]
    \|\mathbf{p}^W_i - \mathbf{R}^W_I \mathbf{p}^I_U - \mathbf{p}^W_I \|^2_2\\
    \|\mathbf{p}^W_j - \mathbf{R}^W_I \mathbf{p}^I_U - \mathbf{p}^W_I \|^2_2\\
    \|\mathbf{p}^W_k - \mathbf{R}^W_I \mathbf{p}^I_U - \mathbf{p}^W_I \|^2_2
    \end{bmatrix}.
    \label{eqn:mul_anchor_meas_eq}
\end{equation}

To prove the state is locally observable, we need to determine the rank of the observability matrix $\mathcal{O}$~\cite{Hermann1977} associated with the system (\ref{eqn:system_dynamics_affine})-(\ref{eqn:mul_anchor_meas_eq}). The matrix $\mathcal{O}$ is constructed by taking the Lie derivatives \cite{Hermann1977} of (\ref{eqn:mul_anchor_meas_eq}) along the system dynamics (\ref{eqn:system_dynamics_affine}). Informally, a Lie derivative computes the change in the system output for a change in the system state. Thus, $\mathcal{O}$ maps changes in state to changes in output. If the matrix $\mathcal{O}$ has a full column rank, then it can be inverted and used to map changes in the output to changes in the state. In the following theorem, we state the conditions under which the system (\ref{eqn:system_dynamics_affine})-(\ref{eqn:mul_anchor_meas_eq}) is locally weakly observable. 

\begin{theorem}
The system with motion model (\ref{eqn:system_dynamics_affine}) and observation model (\ref{eqn:mul_anchor_meas_eq}) is locally weakly observable if:
\begin{description}
\item[(C1)] at least three non-collinear anchors are available;
\item[(C2)] the mobile radio is non-coplanar with the three non-collinear anchors;
\item[(C3)] all three of $a_x, a_y$, and $a_z$ are excited; and
\item[(C4)] all three of $\omega_x, \omega_y$, and $\omega_z$ are excited.
\end{description}
\label{lem:obs_cond}
\end{theorem}
\begin{proof} Due to space constraints,
we provide a sketch of the proof here. The full
proof can be found in our supplementary material\footnote{\url{http://tiny.cc/observability}\label{foot:proof}}. We follow these steps in the proof:

\begin{itemize}
\item[S1.] Calculate the gradients of increasing order Lie derivatives of (\ref{eqn:mul_anchor_meas_eq}) along $f_0, f_1$ and, $f_2$;
\item[S2.] stack the gradients to construct the matrix $\mathcal{O}$;
\item[S3.] perform elementary row and column operations to ensure parts of $\mathcal{O}$ are block-diagonal;
\item[S4.] perform block Gaussian elimination and identify the conditions needed for each block to have a full rank. 
\end{itemize}\qedhere
\end{proof}

Note that these conditions are sufficient conditions. Due to the nonlinear nature of the measurement model (\ref{eqn:mul_anchor_meas_eq}), the matrix $\mathcal{O}$ is dense. Step S3 ensures that parts of the matrix are sparse, thereby facilitating block Gaussian elimination. A key observation in performing step S4 is that the matrix:
\begin{align}
\Delta \mathbf{p}_{ijk} = \begin{bmatrix}[1.2]
    (\mathbf{p}^W_i - \mathbf{R}^W_I \mathbf{p}^I_U - \mathbf{p}^W_I)^T\\
    (\mathbf{p}^W_j - \mathbf{R}^W_I \mathbf{p}^I_U - \mathbf{p}^W_I)^T\\
    (\mathbf{p}^W_k - \mathbf{R}^W_I \mathbf{p}^I_U - \mathbf{p}^W_I)^T
    \end{bmatrix}^T,
    \label{eqn:obs_cond}
\end{align}
is a multiplicative factor of many blocks in $\mathcal{O}$. Hence, ensuring $\Delta \mathbf{p}_{ijk}$ is full-rank is crucial in our analysis. The matrix $\Delta \mathbf{p}_{ijk}$ is full-rank when conditions C1 and C2 are satisfied. Conditions C3 and C4 ensure that the effect of control inputs on the system output is captured in the matrix $\mathcal{O}$ via $f_1$ and $f_2$, respectively. 

\section{Estimation} \label{sec:estimation}
The analysis in the previous section shows that, under certain conditions, the state (\ref{eqn:state}) can be reliably estimated. However, the design or choice of a suitable state estimator cannot be derived from such an analysis. In this section, we present an approach that can be used for spatio-temporal calibration of a tightly-coupled UWB-IMU system. The choice of the estimator is motivated by its applicability to nonlinear systems, computational efficiency, and ability to operate with sparse sensor data.




To simultaneously estimate the state and calibrate the spatio-temporal offset parameters, we use the ESKF \cite{Roumeliotis1999}. In this formulation, inertial dead reckoning is used to propagate the state (\ref{eqn:state}) forward in time using the model (\ref{eqn:system_dynamics_affine}) and the IMU measurements as inputs. The uncertainty associated with dead reckoning is estimated in the prediction step. In the correction step, the error and its uncertainty are estimated by fusing the dead reckoned state and the UWB measurements using their respective uncertainties. The error state corresponding to the state (\ref{eqn:state}) is:
\begin{equation}
    \delta \mathbf{x} = (\delta \mathbf{p}^W_I, \delta \mathbf{v}^W_I, \delta \bm{\theta}^W_I, \delta \mathbf{b}_a, \delta \mathbf{b}_{\omega}, \delta \mathbf{p}^I_U, \delta t_d),
    \label{eqn:error_state}
\end{equation}
where $\delta \mathbf{p}^W_I$ and $\delta \mathbf{v}^W_I$ represent the error in IMU position and translational velocity, respectively. The errors in accelerometer bias, gyroscope bias, spatial offset and temporal offset are represented by $\delta \mathbf{b}_a$, $\delta \mathbf{b}_{\omega}$, $\delta \mathbf{p}^I_U$, and $\delta t_d$ respectively. For a vector-valued parameter such as $\mathbf{p}^I_U \in \mathbb{R}^3 $, the error $\delta \mathbf{p}^I_U$ corresponds to error in individual components: $\delta \mathbf{p}^I_U = [\delta p^I_{Ux}, \delta p^I_{Uy}, \delta p^I_{Uz}]^T$. The local angular error $\delta \bm{\theta}^W_I$ is related to small differential rotations $\delta \mathbf{q}^W_I$ as: $ \delta \mathbf{q} = \left(1, ~\frac{1}{2} \delta \bm{\theta} \right)$, $|\delta \bm{\theta}| \ll  1 $.
Since the accelerometer and gyroscope measurements are used as inputs, the process noise covariance matrix $\mathbf{Q}_p$ for the prediction step is composed of the covariances of the noise and bias terms as defined in Section \ref{sec:imu_model}: $\mathbf{Q}_p \triangleq \text{diag}\left( [\mathbf{Q}_a, \mathbf{Q}_\omega, \mathbf{Q}_{ba}, \mathbf{Q}_{b\omega} ] \right)$. The correction step estimates the error between the dead reckoned state and the state consistent with the measurements. The measurement model for the correction phase of the ESKF is given by (\ref{eqn:meas_model}). The estimated error (\ref{eqn:error_state}) is then composed with the state to compensate for the accumulated drift: $\mathbf{x} = \mathbf{x} \oplus {\delta \mathbf{x}}$, where $\oplus$ is a generic composition operator which represents \emph{(i)} quaternion multiplication for the orientation error, $\mathbf{q}^W_I = \mathbf{q}^W_I \otimes \delta \mathbf{q}$, and \emph{(ii)} the addition operation for the remaining states, $\mathbf{p}^W_I = \mathbf{p}^W_I + \delta \mathbf{p}^W_I$. 
We refer the reader to \cite{Sola2017} for a description of the prediction and correction step of the ESKF. Covariance values for the process noise matrix $\mathbf{Q}_p$ and the measurement noise matrix $Q_r$ were determined using the procedure outlined in Section \ref{sec:real_exp}.
\begin{figure}[!t]
     \centering
     \hspace*{-0.7cm}\input{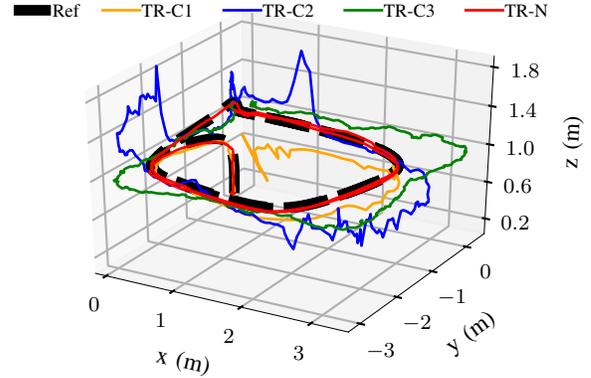}
     \caption{Position estimation results for multiple scenarios: in TR-C1, two anchors are used; in TR-C2 height of mobile UWB radio is at the same level as that of the anchor; in TR-C3, a constant velocity trajectory is commanded; in TR-N, all conditions of Theorem \ref{lem:obs_cond} are satisfied. In all of the experiments, the quadrotor was controlled using ground truth data, the plots show the position output from the estimator. The corresponding RMSE values are provided in Table \ref{tab:rmse}.}
   \label{fig:traj_plot}
\end{figure}
\section{Experiments}
In this section, we present results from simulation and real-world experiments that motivate the need for calibration, validate the theory presented in Section \ref{sec:obs_ana}, and show the stability and accuracy of the estimation approach presented in Section \ref{sec:estimation}. 
To evaluate the accuracy of the proposed approach, position and rotation root-mean-square errors (RMSE) are used as metrics. The rotation RMSE is the Euclidean distance between the Euler angles as defined in\cite{huynh2009}.
\subsection{Simulation experiments}
Our simulation environment is based on Gazebo \cite{gazebo} and uses the Astec Firefly quadrotor from the RotorS  simulator \cite{Furrer2016} as its mobile system. The quadrotor is equipped with a UWB radio and an IMU. Anchors are simulated by fixed radios. Range measurements from UWB radios are perturbed with zero-mean AWGN having a 2\,cm standard deviation. The choice of noise parameters is based on the precision of the UWB radios available on the market \cite{humatics}.

To validate the observability and the identifiability conditions outlined in Section \ref{sec:obs_ana}, we performed multiple simulation experiments where we varied \emph{(i)} the number of anchors, \emph{(ii)} the heights of the anchors, and \emph{(iii)} the velocity profiles of the trajectories commanded to the quadrotor. In all of the experiments, ground truth data was used for the control of the quadrotor. The position estimated by the state estimator is plotted in Fig. \ref{fig:traj_plot} and the corresponding error metrics are provided in Table \ref{tab:rmse}. Each experiment violates one condition from Theorem \ref{lem:obs_cond}; experiment TR-C1 violates condition C1, that is, two anchors are used for estimation; in TR-C2, the anchor heights are adjusted so that the trajectory of the mobile radio is coplanar with the anchors; in TR-C3, a constant velocity trajectory with insufficient excitation of accelerometer and gyroscope axes is commanded; in TR-N, all the conditions outlined in Theorem \ref{lem:obs_cond} are satisfied. In TR-C1, due to insufficient anchors, the positioning diverges gradually (see Fig. \ref{fig:traj_plot}); in TR-C2, the third column in the matrix (\ref{eqn:obs_cond}) is zero, which results in poor position estimates along the z-axis; in TR-C3, the estimated yaw diverges, which results in poor positioning in the xy-plane; in TR-N, the estimated trajectory is close to the commanded trajectory (see Table \ref{tab:rmse}).
\setlength{\tabcolsep}{2pt}
\begin{table}[t!]
\centering
\caption{Estimation performance for the experiments associated with Fig. \ref{fig:traj_plot}. The position and the rotation RMSE are calculated by comparing the estimated state with the ground truth state. $\|\delta \mathbf{p}^I_U \|_2$ and $\| \delta t_d \|_2$ denote the RMSE error in spatial and temporal offset, respectively.}
\begin{tabular}{c c c c c}
\toprule
& Pos. RMSE (m) & Rot. RMSE (rad) & $\|\delta \mathbf{p}^I_U \|_2$ (cm) & $\|\delta t_d\|_2$ (ms)\\
\cmidrule(lr){2-2} \cmidrule(lr){3-3} \cmidrule(lr){4-4} \cmidrule(lr){5-5}
TR-C1 & 0.37 & 0.03 & 6.0 & 3\\
TR-C2 & 0.27 & 0.06 & 1.6 & 48\\
TR-C3 & 0.25 & 0.75 & 2.0 & 21\\
\textbf{TR-N} & \textbf{0.05} & \textbf{0.03} & \textbf{0.9} & \textbf{1}\\
\bottomrule
\end{tabular}
\label{tab:rmse}
\end{table}
\begin{figure}[!b]
    \centering
    \vspace*{-0.2cm}
    \hspace*{-0.5cm}\input{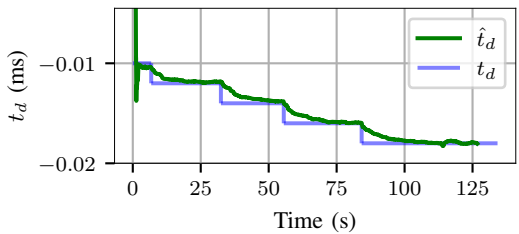}
    \caption{By including a white noise-on-velocity motion model, the estimated temporal offset $\hat{t}_d$ tracks the changing temporal offset $t_d$ reliably.}
    \label{fig:td_var}
\end{figure}
\setlength{\tabcolsep}{5pt}
\begin{table}[b!]
\centering
\caption{Average RMSE from Monte Carlo simulation experiments for different values of spatial and temporal offsets.  $\|\delta \mathbf{p}^I_U \|_2$ and $\| \delta t_d \|_2$ denote the RMSE error in spatial and temporal offset, respectively.}
\begin{tabular}{c c c c}
\toprule
Pos. RMSE (m) & Rot. RMSE (rad) & $\|\delta \mathbf{p}^I_U \|_2$ (cm) & $\| \delta t_d \|_2$ (ms)\\
\cmidrule(lr){1-4}
0.027 & 0.033 & 1.11 & 1.26\\
\bottomrule
\end{tabular}
\label{tab:mc}
\end{table}

To test the stability of the proposed estimation approach, we performed Monte Carlo (MC) simulations varying the magnitude of the UWB-IMU spatial offset between -0.5\,m to 0.5\,m and the temporal offset between -25\,ms to 25\,ms. In all of the MC experiments, the setup was identical to that of TR-N from Fig. \ref{fig:traj_plot}. The results from the MC experiments are provided in Table \ref{tab:mc}. 

The temporal offset may be different from start up to start up of the mobile system or over the duration of its operation. Such situations can be accommodated by using a random walk model: $\dot{t}_d = n_d, ~ n_d \sim \mathcal{N}(0, Q_d)$, where $Q_d$ can be determined empirically from Allan plots \cite{Gyro2006} of the estimated temporal offset. For this experiment, the setup was identical to that of TR-N from Fig. \ref{fig:traj_plot}. The value of $t_d$ was changed dynamically during the simulation. The proposed model manages to estimate varying $t_d$ as shown in Fig. \ref{fig:td_var}.

\begin{figure}[!t]
    \hspace*{-0.5cm}
    \input{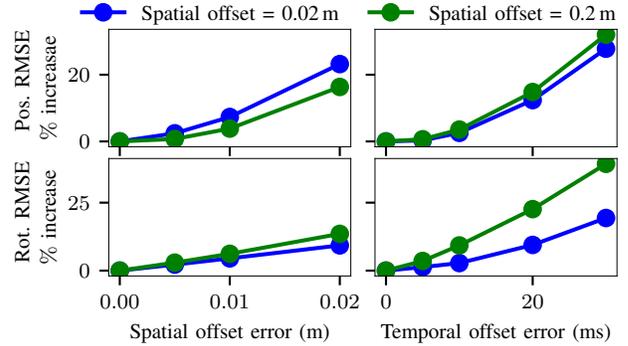}
    \caption{Effect of error in spatial and temporal offset on position and rotation RMSE. The impact of error in spatial and temporal offset is generally more prominent for larger spatial offset ($\| \mathbf{p}^I_U \|_2$ = 0.2\,m) compared to smaller spatial offset ($\| \mathbf{p}^I_U \|_2$ = 0.02\,m).}
\label{fig:param_sens}
\end{figure}

To generally quantify the impact of errors in spatial and temporal offset on the position and the rotation RMSE, we performed multiple experiments by perturbing the parameters and recording the corresponding RMSE values without estimating the spatio-temporal offsets. The results are shown in Fig. \ref{fig:param_sens}: the influence of parametric errors is more prominent for larger spatial offsets. For instance, a 50\% error in the spatial offset increases the position RMSE by 7\% for the smaller baseline ($\| \mathbf{p}^I_U \|_2 = 0.02$m), whereas a 5\% error in the spatial offset for the larger baseline ($\| \mathbf{p}^I_U \|_2 = 0.2$m) increases the position RMSE by 4\%. The effect of the temporal offset is more prominent on the rotation RMSE: an error of 20ms in the temporal offset increases the rotation RMSE by 10\% for the smaller baseline compared to 23\% for the larger baseline. In addition to quantifying the effect of errors in spatio-temporal parameters, these experiments also quantify the expected reduction in position and rotation RMSE after calibration, which supports our discussion of the results from the real-world experiments (see Section \ref{sec:res_dis}).
\subsection{Real-world experiments} \label{sec:real_exp}
\subsubsection{Setup}
\begin{figure}[!t]
     \centering
     \includegraphics[width=0.7\linewidth]{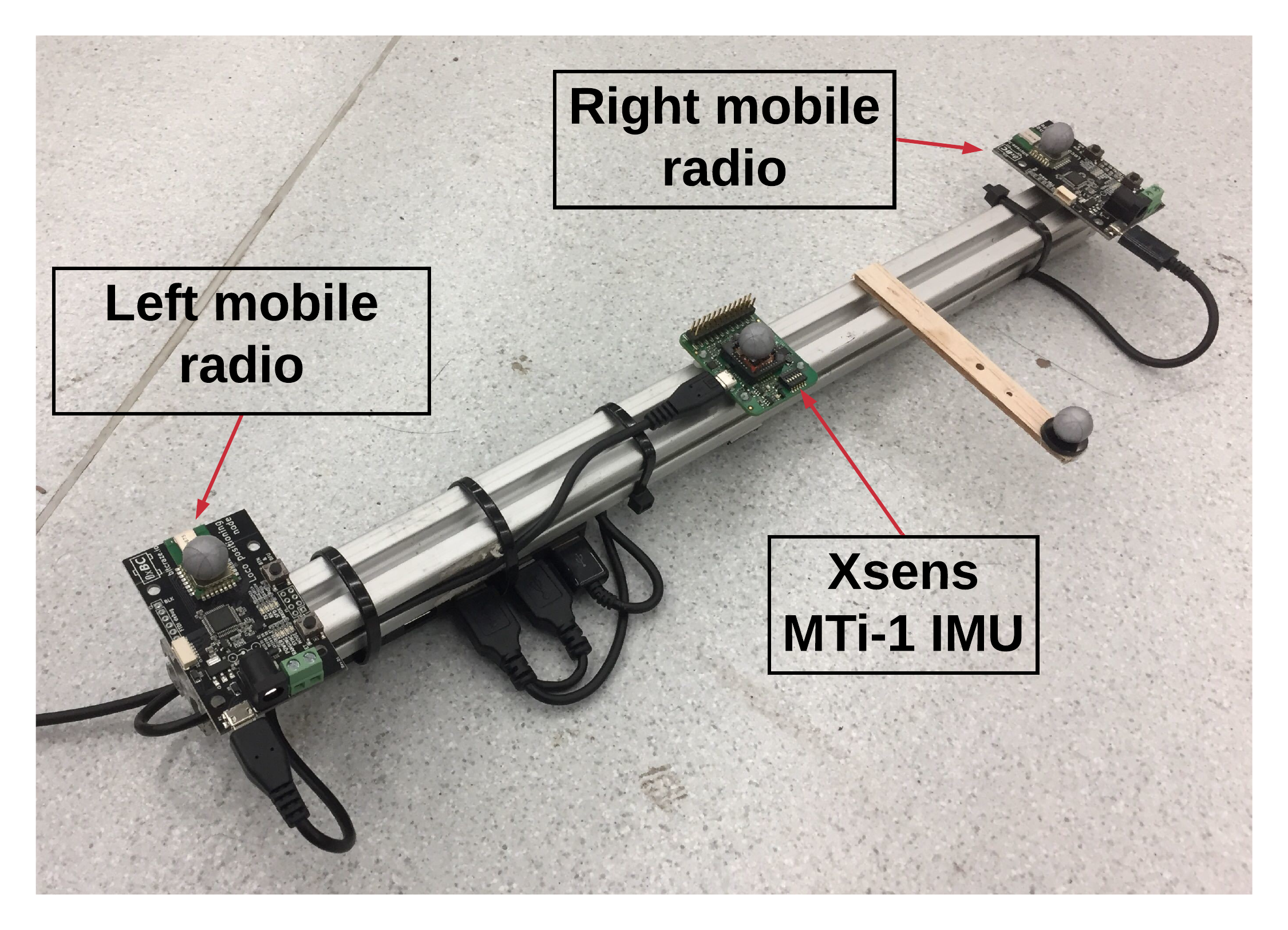}
     \caption{The sensor apparatus consists of a wand equipped with an Xsens MTi 1 IMU and two UWB mobile radios mounted on the left and right side of the IMU. Reflective markers are mounted to obtain ground truth measurements.}
   \label{fig:wand}
\end{figure}

Our setup consists of a sensor wand (see Fig. \ref{fig:wand}), a constellation of 6 anchors, and a motion capture system. Next, we describe the calibration of the anchor positions, followed by a description of the sensor wand.

A benefit of UWB localization systems is that the calibration of anchor positions can be done independently of the mobile system (or wand in our case). Inter-anchor distances can be acquired by configuring one anchor as a receiver and the remaining as transmitters. By choosing one of the anchor’s frame as the origin, the positions of the rest of the anchors can be determined using the inter-anchor distances \cite{batstone2016, hamer2018}. The approach followed here is similar to that of \cite{hamer2018}.  

The sensor wand is equipped with an Xsens IMU and two Decawave UWB radios mounted as shown in Figure \ref{fig:wand}. The IMU yields linear accelerations and angular velocities at 100$\text{Hz}$. The UWB radios operate in TW-ToF mode. Range measurements are acquired at 20$\text{Hz}$ by communicating with the anchors in a round-robin fashion. Ground truth pose information is acquired from a motion capture system.
Parameters for the accelerometer and gyroscope noise models outlined in Section \ref{sec:imu_model} were determined using Allan plots \cite{Gyro2006}. The Decawave UWB radios used in this experiment claim a precision of $\pm$10\,cm. However, different UWB sensors can exhibit different noise characteristics. To estimate $Q_r$, we collected UWB measurements for every mobile-anchor pair by holding the sensor wand static at multiple locations in the operating area. The error in the measured distance was computed by comparing with the ground truth and the variance of the error was used as $Q_r$ in (\ref{eqn:meas_model}).

\subsubsection{Experimental procedure}
At the start of each experiment, the sensor wand was held static for 60s to generate initial estimates for the gyroscope and accelerometer biases. Additionally, the static data were used to generate an initial estimate for roll and pitch values by averaging the \textit{gravity} vector reported by the accelerometer. The position of the IMU was initialized by placing the wand at the origin of the world frame. Note that this is not a precise initialization routine and the representative uncertainty in the initial pose was captured by the initial covariance. The sensor wand was moved manually through various rotation and translation maneuvers in an attempt to simulate sufficient conditions for the state to be locally weakly observable, as per Section \ref{sec:obs_ana}. 
\section{Results and Discussion} \label{sec:res_dis}
We performed multiple experiments to show that joint calibration of spatial and temporal offsets, while simultaneously localizing, is possible when the sufficient conditions outlined in Theorem \ref{lem:td_cond_1} and Theorem \ref{lem:obs_cond} are met. Spatial offsets for both the left and right mobile radio were estimated, due to space constraints, only the results for the left mobile radio are presented here. Results for the right mobile radio are similar. Since the ground truth values for the spatial and the temporal offset were not available, we evaluated the accuracy of the proposed approach by computing the position and rotation RMSE for hand-measured (HM) and self-calibrated (SC) values of the offsets. The HM values for the spatial offsets were obtained by manually measuring the position offset between the IMU and the UWB radio using a ruler. The temporal offset parameter was initialized to zero. 
\begin{figure}[!t]
    \centering
    \hspace*{-0.5cm}\input{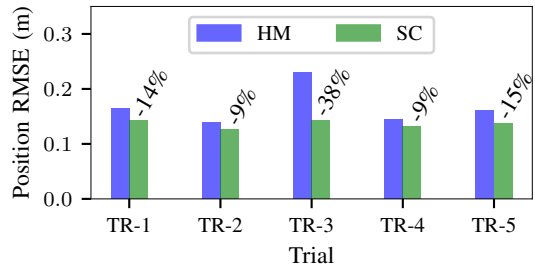}
    \caption{Position RMSE for self-calibrated (SC) and hand-measured (HM) values of the spatial offset and the temporal offset parameters.}
    \label{fig:pos_rmse}
\end{figure}
\begin{figure}[!t]
    \centering
    \vspace*{-0.5cm}
    \hspace*{-0.5cm} \input{figs/rot_rmse_up.pgf}
    \caption{Rotation RMSE for self-calibrated (SC) and hand-measured (HM) values of the spatial offset and the temporal offset parameters.}
    \label{fig:rot_rmse}
\end{figure}
\begin{figure*}[!t]
\centering
    \hspace*{-1cm}\input{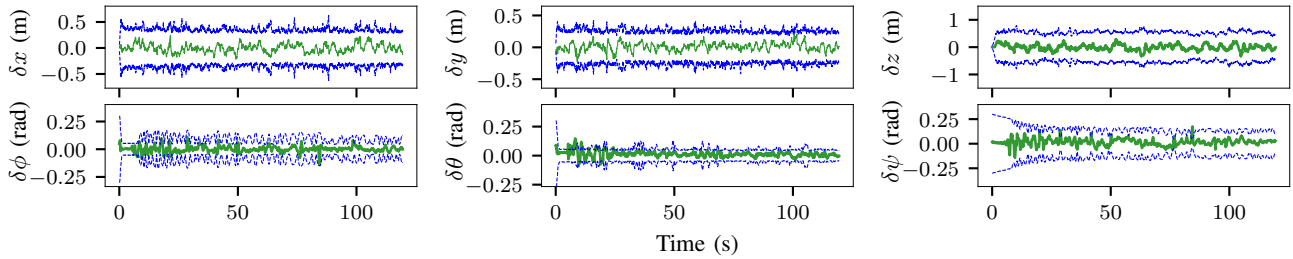}
    \caption{Position error ($\delta x$, $\delta y$ and $\delta z$) plots from experiment TR-1 for SC values of spatial and temporal offset parameters, along with the 3$\sigma$ bounds, are shown in the top row. The corresponding rotation error plots in terms of roll ($\delta \phi$), pitch ($\delta \theta$) and yaw ($\delta \psi$) error are shown in the bottom row. The position and the rotation errors were calculated by comparing the estimated pose from the ESKF with the pose from the motion capture system. Note that the estimation is consistent.}
    \label{fig:pose_err_plot}
\end{figure*}
\setlength{\tabcolsep}{8pt}
\begin{table*}[t!]
\centering
\caption{Estimated values of spatial and temporal offset parameters from 5 experiments (TR-1 to TR-5) along with the 3$\sigma$ uncertainty values. For each experiment, the sensor wand was moved along a different trajectory, ensuring all of the accelerometer and the gyroscope axes were excited. To quantify the drift in the temporal offset over time, experiments TR-3 to TR-5 were performed after the sensor wand had been operating for 4 hours. The temporal offset drifts by $\sim$9\,ms after 4 hours of operation.}
\begin{tabular}{c c c c c c c}
\toprule
Parameter & HM & TR-1 & TR-2 & TR-3 & TR-4 & TR-5\\
\cmidrule(lr){1-1} \cmidrule(lr){2-2} \cmidrule(lr){3-3} \cmidrule(lr){4-4} \cmidrule(lr){5-5} \cmidrule(lr){6-6} \cmidrule(lr){7-7}
$p^I_{Ux} \pm 3\sigma$ (cm) & 2.0 $\pm$ 6.0 & 2.3 $\pm$ 3.4 & 1.6 $\pm$ 3.3 & 1.9 $\pm$ 3.3 & 2.1 $\pm$ 3.5 & 1.6 $\pm$ 3.4\\
$p^I_{Uy} \pm 3\sigma$ (cm) & 19.0 $\pm$ 15.0 & 20.3 $\pm$ 4.2 & 20.1 $\pm$ 3.9 & 20.7 $\pm$ 4.0 & 20.1 $\pm$ 4.7 & 20.4 $\pm$ 4.1 \\
$p^I_{Uz} \pm 3\sigma$ (cm) & 0.0 $\pm$ 0.15 & -0.3 $\pm$ 1.4 & -0.1 $\pm$ 1.4 & -0.2 $\pm$ 1.4 & 0.3 $\pm$ 1.4 & -0.2 $\pm$ 1.4 \\
$t_d \pm 3\sigma$ (ms) & 0.0 $\pm$ 300 & 36.9 $\pm$ 23.9 & 32.9 $\pm$ 24.1 & 44.8 $\pm$ 24.7 & 44.0 $\pm$ 26.4 & 42.4 $\pm$ 28.5 \\
\bottomrule
\end{tabular}
\label{tab:lev_arm_est_res}
\end{table*}
For each experiment (TR-1 to TR-5) the sensor wand was moved manually along a different trajectory. The calibration results are shown in Table \ref{tab:lev_arm_est_res}. Error in the estimated position and rotation was calculated by comparing the estimated pose from the ESKF with the ground truth pose from the motion capture system. The position and rotation error plots for SC parameter values for TR-1, along with the 3$\sigma$ covariance envelopes, are shown in Fig.~\ref{fig:pose_err_plot}. The reduction in the position and the rotation RMSE for SC offset values compared to HM values is shown in Fig. \ref{fig:pos_rmse} and Fig. \ref{fig:rot_rmse}, respectively. To quantify the drift in the temporal offset over time, experiments TR-3 to TR-5 were performed after the sensor wand had been operating for 4 hours. Table \ref{tab:lev_arm_est_res} shows that the temporal offset drifts by $\sim$9\,ms after 4 hours of operation. These results are promising, especially because the precision of the UWB measurements ($\pm$10\,cm) is similar in magnitude to the spatial offset being estimated ($\sim$20\,cm). Furthermore, the reduction in RMSE values agrees with the results from simulation experiments. For a 20\,cm baseline, Fig. \ref{fig:param_sens} shows that a 2\,cm error in spatial offset increases the position RMSE by 18\% and a 30\,ms error in temporal offset increases rotation RMSE by 40\%. The results from Fig. \ref{fig:pos_rmse} and Fig. \ref{fig:rot_rmse} show an average reduction of $\sim$17\% in position RMSE for an improvement of 2\,cm in the spatial offset and an average $\sim$43\% reduction in rotation RMSE for an improvement of 35\,ms in the temporal offset.

\section{Conclusion and Future Work}
In this paper, we derived the conditions for \emph{(i)} the local weak observability of the spatial offset and \emph{(ii)} the local identifiability of the temporal offset of a tightly-coupled UWB-IMU system. An online calibration approach, based on ESKF was proposed.
The results from both simulation and real-world experiments show that if the observability and the identifiability conditions are met, it is possible to accurately calibrate the spatial offset and the temporal offset while simultaneously localizing---without additional sensors or hardware---thus precluding the need for a separate calibration procedure. The addition of a camera to the UWB-IMU system may offer many advantages, as UWB radios and cameras have complementary characteristics. The spatio-temporal calibration of such a system can be seen as a natural extension of the work presented in this paper.
%
\Urlmuskip=0mu plus 1mu\relax
\bibliographystyle{unsrt}
\bibliography{./bib/main.bib}

\begin{thebibliography}{10}

\bibitem{Hol2009}
J.~D. {Hol}, F.~{Dijkstra}, H.~{Luinge}, and T.~B. {Schon}.
\newblock {Tightly coupled UWB/IMU pose estimation}.
\newblock In {\em Proc. of the IEEE International Conference on
  Ultra-Wideband}, pages 688--692, 2009.

\bibitem{Prorok2014}
Amanda Prorok and Alcherio Martinoli.
\newblock {Accurate indoor localization with ultra-wideband using spatial
  models and collaboration}.
\newblock {\em The International Journal of Robotics Research}, 33(4):547--568,
  2014.

\bibitem{Mueller2015}
Mark~W. Mueller, Michael Hamer, and Raffaello D'Andrea.
\newblock Fusing ultra-wideband range measurements with accelerometers and rate
  gyroscopes for quadrocopter state estimation.
\newblock In {\em Proc. of the IEEE International Conference on Robotics and
  Automation (ICRA)}, pages 1--6, 2015.

\bibitem{kaplan2005}
C.~{Hegarty} and E.~{Kaplan}.
\newblock {\em Understanding GPS Principles and Applications, Second Edition}.
\newblock 2005.

\bibitem{Hermann1977}
Robert Hermann and Arthur~J Krener.
\newblock {Nonlinear controllability and observability}.
\newblock {\em IEEE Transactions on Automatic Control}, (5), 1977.

\bibitem{anguelova2008}
Milena Anguelova and Bernt Wennberg.
\newblock State elimination and identifiability of the delay parameter for
  nonlinear time-delay systems.
\newblock {\em Automatica}, 44(5):1373--1378, 2008.

\bibitem{Prorok2011}
Amanda Prorok, Phillip Tom{\'{e}}, and Alcherio Martinoli.
\newblock {Accommodation of NLOS for ultra-wideband TDoA localization in
  single- and multi-robot systems}.
\newblock {\em International Conference on Indoor Positioning and Indoor
  Navigation (IPIN)}, 2011.

\bibitem{Fang2018}
Xu~Fang, Chen Wang, Thien-Minh Nguyen, and Lihua Xie.
\newblock {Graph optimization approach to localization with range
  measurements}.
\newblock {\em arXiv preprint arXiv:1802.10276}, 2018.

\bibitem{Hausman2016}
K.~{Hausman}, S.~{Weiss}, R.~{Brockers}, L.~{Matthies}, and G.~S. {Sukhatme}.
\newblock {Self-calibrating multi-sensor fusion with probabilistic measurement
  validation for seamless sensor switching on a UAV}.
\newblock In {\em Proc. of the IEEE International Conference on Robotics and
  Automation (ICRA)}, pages 4289--4296, 2016.

\bibitem{Kelly2011}
Jonathan Kelly and Gaurav~S. Sukhatme.
\newblock {Visual-Inertial Sensor Fusion: Localization, Mapping and
  Sensor-to-Sensor Self-calibration}.
\newblock {\em The International Journal of Robotics Research}, 30(1):56--79,
  2011.

\bibitem{Weiss2012}
Stephan~M Weiss.
\newblock {\em Vision based navigation for micro helicopters}.
\newblock PhD thesis, ETH Zurich, 2012.

\bibitem{Hesch2014}
Joel~A. Hesch, Dimitrios~G. Kottas, Sean~L. Bowman, and Stergios~I.
  Roumeliotis.
\newblock {Camera-IMU-based localization: observability analysis and
  consistency improvement}.
\newblock {\em The International Journal of Robotics Research}, 33(1):182--201,
  2014.

\bibitem{Li2014}
Mingyang Li and Anastasios~I. Mourikis.
\newblock {Online temporal calibration for camera–IMU systems: Theory and
  algorithms}.
\newblock {\em The International Journal of Robotics Research}, 33(7):947--964,
  2014.

\bibitem{Furgale2013}
P.~{Furgale}, J.~{Rehder}, and R.~{Siegwart}.
\newblock {Unified temporal and spatial calibration for multi-sensor systems}.
\newblock In {\em Proc. of the IEEE/RSJ International Conference on Intelligent
  Robots and Systems (IROS)}, pages 1280--1286, 2013.

\bibitem{Hong2005}
{Sinpyo Hong}, {Man Hyung Lee}, {Ho-Hwan Chun}, {Sun-Hong Kwon}, and J.~L.
  {Speyer}.
\newblock {Observability of error states in GPS/INS integration}.
\newblock {\em IEEE Transactions on Vehicular Technology}, 54(2):731--743,
  March 2005.

\bibitem{skog2011}
I.~{Skog} and P.~{Handel}.
\newblock Time synchronization errors in loosely coupled gps-aided inertial
  navigation systems.
\newblock {\em IEEE Transactions on Intelligent Transportation Systems},
  12(4):1014--1023, 2011.

\bibitem{Roumeliotis1999}
Stergios~I. Roumeliotis, Gaurav~S. Sukhatme, and George~A. Bekey.
\newblock {Circumventing dynamic modeling: evaluation of the error-state Kalman
  filter applied to mobile robot localization}.
\newblock In {\em Proc. of the IEEE International Conference on Robotics and
  Automation (ICRA)}, pages 1656--1663 vol.2, 1999.

\bibitem{Sola2017}
J.~Sol{\`a}.
\newblock {Quaternion kinematics for the error-state Kalman filter}.
\newblock {\em ArXiv}, abs/1711.02508, 2017.

\bibitem{huynh2009}
D.~Huynh.
\newblock {Metrics for 3D rotations: comparison and analysis}.
\newblock {\em Journal of Mathematical Imaging and Vision}, 35:155--164, 2009.

\bibitem{gazebo}
N.~{Koenig} and A.~{Howard}.
\newblock {Design and use paradigms for Gazebo, an open-source multi-robot
  simulator}.
\newblock In {\em 2004 IEEE/RSJ International Conference on Intelligent Robots
  and Systems (IROS) (IEEE Cat. No.04CH37566)}, volume~3, pages 2149--2154
  vol.3, 2004.

\bibitem{Furrer2016}
Fadri Furrer, Michael Burri, Markus Achtelik, and Roland Siegwart.
\newblock {\em Robot Operating System (ROS): The Complete Reference (Volume
  1)}, chapter RotorS---A Modular Gazebo MAV Simulator Framework, pages
  595--625.
\newblock Springer International Publishing, Cham, 2016.

\bibitem{humatics}
{Humatics Rail Navigation System datasheet}.
\newblock
  \url{https://humatics.com/wp-content/uploads/2020/07/Humatics_RailNav_Datasheet_072020.pdf}.

\bibitem{Gyro2006}
{IEEE Std 627-2006 IEEE Standard Specification Format Guide and Test Procedure
  for Single-Axis Laser Gyros, Annex C. IEEE,}.
\newblock 2006.

\bibitem{batstone2016}
K.~{Batstone}, M.~{Oskarsson}, and K.~{Åström}.
\newblock Robust time-of-arrival self calibration and indoor localization using
  wi-fi round-trip time measurements.
\newblock In {\em 2016 IEEE International Conference on Communications
  Workshops (ICC)}, pages 26--31, 2016.

\bibitem{hamer2018}
M.~{Hamer} and R.~{D’Andrea}.
\newblock Self-calibrating ultra-wideband network supporting multi-robot
  localization.
\newblock {\em IEEE Access}, 6:22292--22304, 2018.

\end{thebibliography}
\end{document}